\documentclass[11pt,a4paper]{article}
\usepackage[hyperref]{acl2019}
\usepackage{times}
\usepackage{latexsym}
\usepackage{url}
\usepackage{helvet}  
\usepackage{courier}  
\usepackage{graphicx}  
\usepackage[utf8]{inputenc}
\usepackage{caption}
\usepackage{CJK}
\usepackage{multirow}
\usepackage{amsmath}
\usepackage{amssymb}
\usepackage{color}
\usepackage[title]{appendix}
\usepackage{subcaption}

\usepackage{soul}
\usepackage{booktabs}
\usepackage{algorithm}
\usepackage{algorithmic}
\usepackage{bbm}
\usepackage{setspace}
\usepackage{amsfonts}

\aclfinalcopy 
\def\aclpaperid{42} 

\title{Unsupervised Summarization by Jointly Extracting Sentences and Keywords}

\author{
Zongyi Li$^{1}$, Xiaoqing Zheng$^{1}$, Jun He$^{2}$ \\
$^1$School of Computer Science, Fudan University, Shanghai, China \\
$^2$the Administrative Center of Shanghai R\&D Public Service Platforms \\
{\tt \{zongyili19, zhengxq\}@fudan.edu.cn} \\
{\tt jhe@sgst.cn} \\
}

\begin{document}

\maketitle

\begin{abstract}
We present RepRank, an unsupervised graph-based ranking model for extractive multi-document summarization in which the similarity between words, sentences, and word-to-sentence can be estimated by the distances between their vector representations in a unified vector space. In order to obtain desirable representations, we propose a self-attention based learning method that represent a sentence by the weighted sum of its word embeddings, and the weights are concentrated to those words hopefully better reflecting the content of a document. We show that salient sentences and keywords can be extracted in a joint and mutual reinforcement process using our learned representations, and prove that this process always converges to a unique solution leading to improvement in performance. A variant of absorbing random walk and the corresponding sampling-based algorithm are also described to avoid redundancy and increase diversity in the summaries. Experiment results with multiple benchmark datasets show that RepRank achieved the best or comparable performance in ROUGE.
\end{abstract}
\section{Introduction}

Automatic document summarization is one of big natural language processing (NLP) challenges in a world of information overload. A good summary should preserve the most important content of a text and also present the content in well syntactic structure. Summarization techniques can be divided into two predominant categories: extractive and abstractive. Extractive methods aim to select salient sentences from a document and stitch them together to produce in a condensed version, while abstractive ones attempt to concisely paraphrase the important content in the document.

Neural sequence-to-sequence techniques for abstractive document summarization  \cite{emnlp-15:Rush,naacl-16:Chopra,emnlp-16:Takase,emnlp-17:Li-2,acl-17:Tan,acl-17:Zhou,acl-17:See} are becoming increasingly popular thanks to the success of generative neural models for text \cite{iclr-15:Bahdanau}. However, extractive methods are still attractive as they are less computational expensive and less depending on the size of manually prepared dataset. Besides, they are usually able to produce grammatically and semantically coherent summaries \cite{acl-15:Schluter,naacl-15:Li,acl-16:Peyrard,acl-16:Cheng,emnlp-17:Wang,aaai-17:Nallapati}.

Extractive summarization can be cast as a budgeted subset selection problem \cite{ecal-07:McDonald,acl-11:Lin} where a document is considered as a set of sentences and the summarization is to select a subset of the sentences under a length constraint. Global optimization methods such as integer linear programming (ILP) have been demonstrated to be powerful for solving this discrete optimization problem by maximally covering important language concepts (e.g. words, phrases and sentences) in a resulting summary \cite{tac-09:Gillick,acl-13:Li,naacl-15:Li,acl-16:Durrett}.

In ILP-based summarization systems, the objective is normally formalized as a weighted sum of the language concepts, and how to choose the concepts and determine their weights does matter for those systems. The parameters for the concept selection and weight estimation need to be trained to achieve competitive performance. However, the values of parameters learned from a data set might not be optimal for another, especially for the those from different domains, and such data-driven methods cannot be applied when the training data is unavailable. Graph-based ranking methods \cite{emnlp-04:Mihalcea,naacl-07:Zhu,emnlp-17:Wang} seem to be more appealing in such situations, in which the language concepts can be ranked automatically using a variant of random walk algorithm \cite{tr-99:Page}.

When the graph-based ranking methods are applied, the process for ranking the language concepts begins by expressing a directed weighted graph as $n \times n$ ``transition matrix'', $P$, where $n$ is the number of concepts. For text summarization, each element $P_{ij}$ represents the likelihood that a random surfer moves to $j$ from $i$, which is normally derived from the similarity score between the language concept $i$ and $j$. Term frequency-inverse document frequency (tf-idf) is often used to calculate such similarity scores \cite{acl-07:Wan,emnlp-17:Wang}. The tf-idf weighting scheme can be successfully used for ranking the relevance among documents, but might not be well applicable to scoring the similarities between words or sentences (smaller language units). Besides, the if-idf does not capture sentences which are semantically equivalent but expressed with different words.

Recently, word embeddings has been empirically proven to be quite useful for various NLP tasks \cite{mikolov2013efficient,pennington2014glove}, in which words are mapped to dense vectors in a low-dimensional embedded space, and these word vectors keep meaningful linguistic characteristics that words sharing similar meanings aggregate together whereas dissimilar words repel each other. If sentence representations can be properly derived from their word's embeddings, all the sentences and words are embedded into a shared vector space, and then the similarity of sentence-to-sentence and sentence-to-word would be better estimated by simply measuring the distance in this space. We explore such possibility by developing a method to learn such representations based on self-attention mechanism, and seeing whether it can further improve the graph-based ranking summarization.

The graph-based ranking methods generates a summary by ranking all the sentences, and taking the top ones. A good sentence should be a representative of many other similar sentences, so that it likely conveys the central meaning of a document. However, such methods tend to select multiple near-identical sentences. In order to make the top sentences different from each other, Zhu et al \shortcite{naacl-07:Zhu} suggested that the next sentence is chosen with the largest expected number of visits before absorption in random walk after the selected sentence are turn into absorbing states. This solution does increase diversity in the summary, but it may select dull sentences like ``it is alright.'' due to its selection strategy (such sentences are semantically ``isolated'' from others, thus may have the largest expected steps to the absorbing states). We adopt the idea of absorbing random walks, but use a sampling-based algorithm to select the next sentences to avoid such problem.




\section{Graph-based Summarization Model}


In this section, we describe how to extract the summary sentences and keywords from a single or multiple documents in a joint and unsupervised way. The sentences and words can be ranked and extracted iteratively with mutual reinforcement. We prove that such process always comes to converge and has a unique solution. We also propose a sentence representation learning method specially designed for the summarization task. In this method, a sentence is represented by the weighted sum of its word representations, and the weights are concentrated to those ``keywords'' better reflecting the content of a document. Finally, a sampling algorithm for random walks in an absorbing Markov chain is presented to reduce the redundancy for the generated summaries.

\subsection{Similarity based on Embeddings}

When a set of sentences is ranked by teleporting random walks, we want to make sure that the important words (or keywords) are covered by the top-ranked sentences. Inspired by \cite{sigir-02:Lin,acl-07:Wan}, we make that happen by simultaneously ranking the sentences and words in a mutually reinforcing way. A salient sentence should be ``recommended'' by many other sentences and also be ``voted'' by the salient words due to their common ``interest'' in expressing the similar meaning, and a word is important if it occurs in many salient sentences and heavily linked with other keywords.

RepRank requires a weight matrix $P^s$, a word-to-sentence similarity matrix $M^s$, a probability distribution $V^s$ that encodes a pre-specified ranking as prior knowledge for ranking sentences. $P^s$ is a $n \times n$ symmetric matrix, where $P^s_{ij}$ is the non-negative weight on the edge from sentence $i$ to $j$ (their similarity between sentence $i$, $j$), and $n$ denotes the number of sentences to be ranked. $M^s$ is a $m \times n$ weight matrix, where $m$ is the number of words (including unigram and bigram), and the element in row $i$ and column $j$ is the similarity score between word $i$ and sentence $j$. The prior ranking is represented as a row vector $V^s = (V^s_1, ..., V^s_n)$ such that $\Sigma^n_{i = 1}V^s_i = 1, V^s_i \ge 0$. The highest priority sentence has the largest probability, the next has smaller one, and so on. For ranking words, a similar $m \times m$ word-to-word matrix $P^w$, a sentence-to-word $n \times m$ similarity matrix $M^w$ (the transpose of $M^s$) and a probability distribution $V^w$ need to be provided.

In order to rank the sentence and words, we first define an $n \times n$ transition matrix $\widetilde{P}^s$ by normalizing the rows of $P^s$: $\widetilde{P}^s_{ij} = P^s_{ij}/\Sigma^n_{k = 1}P^s_{ik}$, so that $\widetilde{P}^s_{ij}$ is viewed as the probability that a walker moves from $i$ to $j$. The normalized $\widetilde{M}^s$ is also obtained by applying the similar transformation over the word-to-sentence matrix $M^s$. We then make a teleporting random walk $\hat{P}^s$ by interpolating each row with the pre-specified prior distribution $V^s$ and the distribution reflecting the votes from words as follows:
\begin{equation}\label{transition_matrix_sentence}  \small
  \hat{P}^s = \alpha \widetilde{P}^s + \beta \mathbbm{1} V^s  + \gamma \mathbbm{1} V^w \widetilde{M}^s
\end{equation}
\noindent where $\mathbbm{1}$ is the column vector of ones, and $\alpha, \beta, \gamma \in [0, 1]$ are the weights that balance the three parts. The sum of $\alpha, \beta$ and $\gamma$ is equal to $1$. The term governed by $\gamma$ represents the impact of the word distribution on ranking sentences. If $\hat{P}^s$ does not have zero elements (note that $V^s$ has no zero elements), this teleporting random walk is irreducible and aperiodic. All the states are positive recurrent and thus ergodic. Therefore $\hat{P}^s$ has a unique stationary distribution $\pi^s = \pi^s \hat{P}^s$, where $\pi^s$ is a row probability distribution vector. We say that $\lambda = 1$ is the dominant eigenvalue of $\hat{P}^s$, and $\pi^s$ is the corresponding dominant left eigenvector of $\hat{P}^s$. The $i^{th}$ entry of $\pi^s$ is the ranked score for sentence $i$. The highest-ranked sentence $s_*$ can be taken by $s_* = \text{argmax}^n_{i = 1} \pi_i^s$. Likewise, we can make a teleporting random walk $\hat{P}^w$ for ranking words:
\begin{equation}\label{transition_matrix_word}  \small
  \hat{P}^w = \alpha \widetilde{P}^w + \beta \mathbbm{1} V^w  + \gamma \mathbbm{1} V^s \widetilde{M}^w
\end{equation}
We are ready to describe how to obtain the matrices of $P^s$, $P^w$, $M^s$ and $M^w$. Assuming that words and sentences can be represented in a shared vector space through unsupervised learning, the cosine distance between any two vectors could be used to estimates how similar they are. Each element of those matrices is simply calculated by:
\begin{equation}\label{cosine distance}  \small
  e_{ij} = \left\{
             \begin{array}{llr}
             \frac{v_i v_j}{||v_i|| \cdot ||v_j||} & \text{if} \ \frac{v_i v_j}{||v_i|| \cdot ||v_j||} > \epsilon \ \text{and} \ i \neq j \\
             0 & \text{otherwise.}
             \end{array}
\right.
\end{equation}
\noindent where $v_i$ and $v_j$ are word or sentence vector representations according to which matrix is calculated. A threshold $\epsilon$ is used to create a sparse graph like \cite{jair-04:Erkan}.



\subsection{Learning Word and Sentence representations with Self-Attention Mechanism}

If words and sentences can be embedded into a shared vector space, and we can measure their similarity by calculating the distances between them in such space. A simple method is to represent a sentence by the weighted sum of its word embeddings, and the remaining problem is how to assign appropriate weights to the words. Inspired by Lin et al \shortcite{lin2017structured}, a self-attention mechanism is used to estimate such weights by allowing each word of a sentence competing against the others when representing the sentence. Sentence representations are trained by better predicting the following sentences in a document. With this training object, the sentences within the same document are linked together mainly by the words with more weights, and those ``winners'' (words) can be viewed as good candidates for keywords.

Given a sentence $s$ that is a sequence of $m$ word $w_i$ ($1 \ge i \ge m$), each word $w_i$ is associated with its $d$-dimensional embedding $e_{w_i}$, and the sentence is represented as follows:
\begin{equation} \small \label{eq:sentence}
e_s = \sum\nolimits_{i=1}^{m} a_i e_{w_i}
\end{equation} 

\noindent where $a_i$ is the weight of word $w_i$, which is computed using the self-attention mechanism as follows:
\begin{equation}	\small
    a = \text{softmax}(W_2 \tanh(W_1 (e_{w_1}, \cdots, e_{w_m})^{\top}))
\end{equation}
\noindent where $W_1$ is a wight matrix with a shape of $h$-by-$d$ ($h$ is a hyper-parameter), and $W_2$ is a vector of parameters with size $h$. The softmax function ensures all the computed weights sum up to $1$.

For a document consisting of $n$ sentences $s_j$ ($1 \le j \le n$), Eq. (\ref{eq:sentence}) is used to map each sentence $s_j$ into its representation $e_{s_j}$. A long short-term memory (LSTM) network summarizes all the preceding $t$ sentences, and produces a context representation $c_t = \text{LSTM}(s_{\leq t})$. Following Oord et al \shortcite{oord2018representation}, given a context $c_t$, $k$ future sentences need to be predicted, and a log-bilinear model is used to make such prediction:
\begin{equation}
    f_k(s_{t + k}, c_t) = \exp(s_{t + k}^{\top} W_k c_t)
\end{equation}
\noindent where $W_k$ is used for the prediction at step $k$. Given a set $S = \{s_1, \cdots, s_z\}$ of $z$ random samples containing one positive sentence $s_{t + k}$, and $(z - 1)$ negative sentences sampled from other documents. A loss based on noise-contrastive estimation \cite{gutmann2010noise} can be written as:
\begin{equation}
    l_k(c_t) = -\mathbb{E}_{S}\bigg( \log 
    \frac{f_{k}(s_{t + k}, c_{t})}
    {\sum_{s_{j} \in S} f_{k}(s_{j}, c_{t})} \bigg)
\end{equation}
\noindent Both the sentence representations and parameters of LSTM are trained by optimizing the above loss. The word embeddings and their weights used to produce the sentence representations are also tuned accordingly.

\subsection{Mutual Reinforcement Process and Convergence}

Intuitively, the stationary distributions of $\hat{P}^s$, $\hat{P}^w$ in Eq. (\ref{transition_matrix_sentence}) and (\ref{transition_matrix_word}) can be solved in an iterative way. Specifically, we first obtain the dominant left eigenvector $\pi^s$ of $\hat{P}^s$ in Eq. (\ref{transition_matrix_sentence}) by $\pi^s = \pi^s \hat{P}^s$, and then compute the stationary distribution $\pi^w$ of $\hat{P}^w$ in Eq. (\ref{transition_matrix_word}) after replacing $V^s$ with $\pi^s$. The obtained distribution $\pi^w$ will be used to replace $V^w$ of Eq. (\ref{transition_matrix_sentence}) to compute new dominant eigenvector of the updated $\hat{P}^s$. The above steps are alternately performed until convergence. In this iterative process, the sentences are ranked by taking into account the word's relative importance and vice versa, and the both tasks are mutually reinforced by allowing the reciprocal information to flow back and forth. Such iterative algorithm seems work well, but the question still remain as to whether the iteration can converge and lead to a unique solution, and if yes, whether there exists a more efficient algorithm to find the same solution.

\begin{figure}[htbp]
 \small
 \centering
 \includegraphics[width=6.0cm]{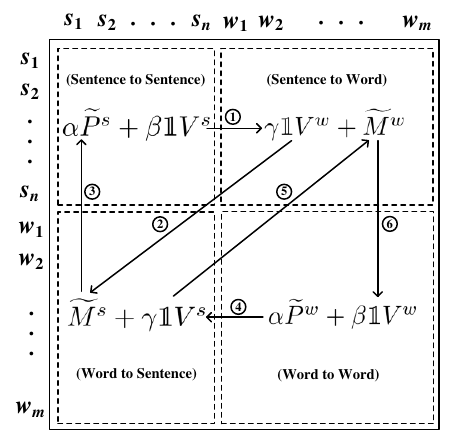}
 \caption{A matrix $P$ that represents a directed graph whose vertices are the sentences ($s_1, ..., s_n$) and words ($w_1, ..., w_m$) from a single or multiple texts. Each element $P_{ij}$ of the matrix represents the probability that a random surfer moves to the vertex $j$ from $i$.}
 \label{fig:graph}
\end{figure}

Imagine a random walker over the graph $\widetilde{P}^s$ of Eq. (\ref{transition_matrix_sentence}). At each step, the walker can do one of three things: she moves to a neighboring sentence according to their similarities with probability $\alpha$; she walks to a random state based on the prior distribution $V^s$ with probability $\beta$; otherwise she is teleported to a word vertex randomly according to the distribution $V^w$, and moves back to any other sentence according to the word-to-sentence weights $\widetilde{M}^s$. We have a similar interpretation for Eq. (\ref{transition_matrix_word}), and so can represent the graph consisting of both the sentences and words as a larger matrix $P$, as shown in Figure \ref{fig:graph}. The terms in the right side of Eq. (\ref{transition_matrix_sentence}) and (\ref{transition_matrix_word}) are assigned into four submatrices accordingly.

The upper left submatrix of $P$ represents the weights that determine the way to linger among the sentences, while the lower right one defines how to move among the words. The sentence-to-word and word-to-sentence walks are randomly specified by the weights of the upper right and lower left submatrices respectively. The rightmost term of Eq. (\ref{transition_matrix_sentence}) is functioned as a three-step moves demonstrated by path $1 \to 2 \to 3$ in Figure \ref{fig:graph}: with probability $\gamma$, she walks to the word vertex in the next step rather than neighbor sentence, then chooses a word to move based on the distribution $V^w$, and finally walks back to a sentence randomly according to the word-to-sentence similarity matrix $\widetilde{M}^w$. Likewise, the path $4 \to 5 \to 6$ demonstrates how the last term in Eq. (\ref{transition_matrix_word}) contributes. Hence, the successive and iterative process for finding the stationary distributions of $\hat{P}^s$ and $\hat{P}^w$ is equal to teleporting random walks on the graph represented by $P$.

If we ensure both the $V^s$ and $V^w$ have no zero element and define a transition matrix $\hat{P}$ by normalizing the rows of $P$, each element $\hat{P}_{ij}$ is guaranteed to be greater than $0$, and the sum of elements in each row of $P$ is equal to $1$. For such row stochastic matrix, $\lambda = 1$ is not a repeated eigenvalue of $\hat{P}$ and is greater in magnitude than any other eigenvalue. Therefore, the eigensystem, $\pi = \pi \hat{P}$, has a unique solution, where $\pi$ is the stationary distribution with the dominant eigenvalue.


If the eigenvalue is known, the problem of eigensystem can be restated as a linear system, and the complexity of computing the corresponding eigenvector is $O(n^3)$ for any $n \times n$ matrix. So the algorithm for obtaining the stationary distribution using the expanded matrix $\hat{P}$ has a complexity of $O((n + m)^3)$, where $n$ denotes the number of sentences, and $m$ the number of words. The complexity of the iterative algorithm is $c(n^3 + m^3)$, where $c$ the number of iterations required for convergence. Assuming $m = an$, we have:
\begin{equation} \label{computational_complexity}  \small
\begin{array}{ll}
    & c(n^3 + m^3) - (n + m)^3 \\
  = & (cn^3 + cm^3) - (m^3 + n^3 + 3nm^2 + 3mn^2) \\
  = & ((c - 1)a^3 - 3a^2 - 3a + c - 1)n^3 \ (\text{by} \ m = an) \\
  \ge & 0 \ (\text{if} \ a \ge 2 \ \text{and} \ c \ge 3)
\end{array}
\end{equation}
\noindent Hence, the random walk algorithm using the expanded matrix $\hat{P}$ is generally quicker and easier, which is used in RepRank.

\subsection{Approximate sampling algorithm for Absorbing Random Walks}

The ranking algorithm based on absorbing random walks can be used to improve diversity and reduce redundancy by making the ranked sentences become absorbing states that ``drag down'' the salience of similar unranked sentences \cite{naacl-07:Zhu}. Once the stationary distributions $\pi^s$ and $\pi^w$ are obtained by the algorithm described above, we compute the updated $\hat{P}^s$ using Eq. (\ref{transition_matrix_sentence}) after replacing the $V^w$ with $\pi^w$. The first sentence $s_k$ with highest score can be picked by $s_k = \text{argmax}^n_{i = 1} \pi_i^s$, and then we turn this sentence into an absorbing state by setting $\hat{P}^s_{kk} = 1$ and $\hat{P}^s_{kj} = 0, \forall j \neq k$. Zhu et al \shortcite{naacl-07:Zhu} proposed to choose the next sentence with the largest expected number of visits before absorption. As mentioned in the introduction, this selection strategy is likely to pick dull sentences since those sentences are semantically ``isolate'' from others and will visit themselves many times via self-edges (note that the diagonal elements are set to $1$ before the row normalization in their implementation).

The idea behind our sampling algorithm is simple, which simulates the process of random walks on the graph with absorbing states. For each sampling process, we choose the first state $i$ randomly sampled from an uniform distribution, and moves to the next state $j$ randomly according to the $i$-th row probability distribution $(\hat{P}^s_{i1}, ..., \hat{P}^s_{in})$. For each iteration $k$, we keep the current estimated distribution, denoted as $\pi^s_{(k)}$. Once such random walk reaches any absorbing state, the walk is absorbed and a sampling process stops. This process will repeat with sufficient times in parallel, and for each state, the number of visits is counted during the process. The counted visits are normalized to produce $\pi^s_{(k + 1)}$. The iteration continues until convergence, which is defined as the Kullback–Leibler divergence from $\pi^s_{(k + 1)}$ to $\pi^s_{(k)}$ being less than a pre-specified threshold $\kappa$, namely $D_{KL}(\pi^s_{(k)}||\pi^s_{(k + 1)}) \le \kappa$. The words can be ranked in the same way. The ranked sentences or words are extracted under a given length constraint. The complete RepRank algorithm is summarized in Figure \ref{fig:algorithm}.

\begin{figure}[htbp]
 \footnotesize

 \begin{spacing}{0.9}

 \rule{240pt}{0.5pt}

 \hangafter 0 \hangindent 0.5em \textbf{Inputs:} multi-document consisting of $n$ sentences and $m$ words.

 \vspace{2.0pt}

 \hangafter 0 \hangindent 0.5em \textbf{Output:} a summary and $d$ keywords.

 \vspace{2.0pt}

 \hangafter 0 \hangindent 0.5em  \textbf{Algorithm:}

 \vspace{2.0pt}

 \hangafter 0 \hangindent 0.5em  $1$: construct a $(n + m) \times (n + m)$ matrix $P$:

 \vspace{0.5pt}

 \hangafter 0 \hangindent 1.5em $\begin{Bmatrix}
                                 \alpha \widetilde{P}^s + \beta \mathbbm{1} V^s & \gamma V^w + \widetilde{M}^w \\
                                 \widetilde{M}^s + \gamma \mathbbm{1} V^s & \alpha \widetilde{P}^w + \beta \mathbbm{1} V^w
                                 \end{Bmatrix}$.

 \vspace{2.0pt}

 \hangafter 0 \hangindent 0.5em  $2$: define a transition matrix $\hat{P}$ by normalizing the rows of $P$.

 \vspace{2.0pt}

 \hangafter 0 \hangindent 0.5em  $3$: compute the stationary distribution or dominant eigenvector

 \vspace{0.5pt}

 \hangafter 0 \hangindent 1.5em       $\pi = (\pi^s, \pi^w)$ by solving the eigensystem $\pi = \pi \hat{P}$.

 \vspace{2.0pt}

 \hangafter 0 \hangindent 0.5em  $4$: set $\widetilde{\pi}^s$: $\widetilde{\pi}^s_{i} = \pi^s_{i}/\Sigma^n_{k = 1}\pi^s_{i}$.

 \vspace{0.5pt}

 \hangafter 0 \hangindent 3.0em           $\widetilde{\pi}^w$: $\widetilde{\pi}^w_{i} = \pi^w_{i}/\Sigma^m_{k = 1}\pi^w_{i}$.

 \vspace{0.5pt}

 \hangafter 0 \hangindent 3.0em           $\hat{P}^s = \alpha \widetilde{P}^s + \beta \mathbbm{1} \widetilde{\pi}^s  + \gamma \mathbbm{1} \widetilde{\pi}^w \widetilde{M}^s$.

 \vspace{0.5pt}

 \hangafter 0 \hangindent 3.0em           $\hat{P}^w = \alpha \widetilde{P}^w + \beta \mathbbm{1} \widetilde{\pi}^w  + \gamma \mathbbm{1} \widetilde{\pi}^s \widetilde{M}^w$.

 \vspace{2.0pt}

 \hangafter 0 \hangindent 0.5em  $5$: \textbf{if} ranking sentences

 \vspace{0.5pt}

 \hangafter 0 \hangindent 2.5em       \textbf{then} $\hat{P} = \hat{P}^s$, $\pi = \widetilde{\pi}^s$.

 \vspace{0.5pt}

 \hangafter 0 \hangindent 2.5em       \textbf{else} \hspace{1.5pt} $\hat{P} = \hat{P}^w$, $\pi = \widetilde{\pi}^w$.

 \vspace{2.0pt}

 \hangafter 0 \hangindent 0.5em  $6$: \textbf{repeat} until all sentences or words are ranked:

 \vspace{2.0pt}

 \hangafter 0 \hangindent 0.5em  $7$: \hspace{2.0pt} select the item $t = \text{argmax}_{i} \pi_i$ from unranked items

 \vspace{2.0pt}

 \hangafter 0 \hangindent 0.5em  $8$: \hspace{2.0pt} turn the selected item $t$ into absorbing state.

 \vspace{2.0pt}

 \hangafter 0 \hangindent 0.5em  $9$: \hspace{2.0pt} update the distribution $\pi$ using the sampling algorithm

 \vspace{0.5pt}

 \hangafter 0 \hangindent 0.5em \hspace{12.0pt} for absorbing random walks.

 \vspace{2.0pt}

 \hangafter 0 \hangindent 0.1em  $10$: generate a summary by extracting the top sentences under the

 \vspace{0.5pt}

 \hangafter 0 \hangindent 0.5em \hspace{8.0pt} length constraint, and take the top $d$ words as keywords.

 \rule{240pt}{0.5pt}

 \end{spacing}

 \caption{\small{The RepRank algorithm.}}
 \label{fig:algorithm}
\end{figure}

\section{Experiment}

We evaluated our graph-based ranking algorithm on several multiple-document summarization (MDS) datasets. In this section, we describe the datasets used to evaluate for MDS tasks, the baseline systems we compare with, implementation details of our model, and the experimental results.

\subsection{Datasets}

The experiments were performed using data from two years of Document Understanding Conference (DUC): $2002$ and $2007$. DUC conducts large scale evaluation of automatic systems on different summarization tasks. These conferences have been held every year since $2001$ and the test sets and evaluation methods adopted by DUC have become the standard for reporting results in publications.

In the early years of DUC, the test set comprised a variety of inputs such as science, business, politics, law, society, and sports, collections of multiple events, opinions, and descriptions of single events. Later years switched to more single-event-type test sets. A collection of documents belonging to the same topic are grouped together into a cluster. Each cluster is associated with $2$ to $4$ reference summaries written by human experts. On DUC $2002$ corpus, following the official guidelines, we used the limited length ROUGE recall metric at $200$ words for the MDS task. DUC $2007$ contain $45$ topics. Each topic has $25$ news documents and $4$ model summaries. The length of the model summary is limited to $250$ tokens.

\subsection{Baseline Systems}

We compare our RepRank with some unsupervised baseline and state-of-the-art systems. We just excerpted the numbers reported in the literature or the results produced by the published tools for fair comparison. The systems compared on the three datasets would be different slightly. The following systems are chosen for comparison: SPARSE \cite{aaai-15:Liu}, ICSI \cite{eacl-14:Hong}, SFOUR \cite{ecal-12:Sipos}, LexRank \cite{jair-04:Erkan}, TextRank \cite{emnlp-04:Mihalcea} and Lead \cite{acl-98:Wasson}. In our experiments, we used the implementation available in the sumy package.

\subsection{Implementation}

We leveraged a large unlabeled corpus, English Wikipedia, to pre-train distributed word representations by GloVe model \cite{pennington2014glove}. Preliminary experiments show that averaging the representations of words (after stopwords being removed) performs better than tf-idf weighted word vectors for generating sentence representations. Thus, we also evaluated a variant of RepRank where sentence representations are produced by simply averaging their word representations to see the impact of our self-attention based learning method.

We construct an initial ranking for each sentence by calculating $\rho^{-\tau}$ like \cite{naacl-07:Zhu}, where $\rho$ is the position of the sentence in its text, and $\tau$ is a hyper-parameter that is set to $0.25$ in our implementation. For MDS task, we normalize over all sentences in all texts to form a distribution that gives higher probability to sentences closer to the beginning. With a larger $\tau$, the probability assigned to later sentences decays more rapidly. A frequency distribution is taken as the prior ranking for the words in the text.

In order to create a relatively sparse graph, we use the cosine threshold values of $0.45$, $0.3$, and $0.2$ for computing the sentence-to-sentence, sentence-to-word, word-to-word similarities in Eq. (\ref{cosine distance}), which are chosen slightly above their average cosine distances. The parameters $\alpha$, $\beta$ and $\gamma$ in Eq. (\ref{transition_matrix_sentence}) and (\ref{transition_matrix_word}) are set to $0.70$, $0.05$ and $0.25$ respectively. We set the threshold $\kappa = 0.0001$ as the convergence condition defined by Kullback–Leibler divergence.

\subsection{Evaluation}

A summary that has higher similarity with the source text can be considered better than one with lower similarity. System summaries are commonly evaluated using ROUGE \cite{tsbo-04:Lin}, a recall oriented metric that measures the $n$-gram overlap between a system summary and a set of human-written reference summaries. These overlap scores have been shown to correlate well with human assessment, and so ROUGE removes the need for manual judgments in this part of evaluation. ROUGE scores are computed typically using unigram (R-$1$) or bigram (R-$2$) overlaps. 
In our experiments below, we report the scores using R-$1$ and R-$2$ metrics under the pre-specified length constraints for each dataset.

\begin{table} [htbp] \footnotesize
\setlength{\abovecaptionskip}{-0.0cm}
\setlength{\belowcaptionskip}{0.0cm}
\caption{\label{performance_duc_2002} \small{Comparison with state-of-the-art systems on DUC-$2002$.}}
\begin{center}
\begin{tabular}{p{50mm}|p{9mm}|p{9mm}}
\hline
\hline
{\bf Model} & {\bf R-1} & {\bf R-2} \\
\hline
LexRank \cite{jair-04:Erkan} & $0.447$ & $0.144$ \\
TextRank \cite{emnlp-04:Mihalcea} & $0.455$ & $0.170$ \\
SFOUR \cite{ecal-12:Sipos} & $0.442$ & $0.181$ \\
ICSI \cite{eacl-14:Hong} & $0.445$ & $0.155$ \\
\hline
RepRank (Glove) & $0.479$ & $0.196$  \\
RepRank (Self-attention) & $0.482$ & $0.203$ \\
RepRank (Self-attention) + Absorb & $0.476$ & $0.199$ \\

\hline
\hline

\end{tabular}
\end{center}
\end{table}

\subsection{Results}

We report the experimental results for DUC-$2002$ and $2007$ in Table \ref{performance_duc_2002} and \ref{performance_duc_2007} respectively. The ``Glove'' in the parenthesis indicates that which unsupervised learning algorithm is used to train the word representations. From these numbers, a handful of trends are readily apparent. First, we note that ``full-fledge'' RepRank indicated by ``(self-attention) + Absorb'' is superior to that the ranking algorithm based on absorbing random walks is not used to reduce redundancy and improve diversity in the summary. RepRank augmented with our proposed sentence learning method, indicated by ``(self-attention)'' performs slightly better than that with GloVe because the former is able to conveys both the global statics (topic-related) and local context information into the learned word embeddings which do help RepRank to capture the interactions between words and sentences across domains.



\begin{table} [htbp] \footnotesize
\setlength{\abovecaptionskip}{-0.0cm}
\setlength{\belowcaptionskip}{0.0cm}
\caption{\label{performance_duc_2007} \small{Comparison with state-of-the-art systems on DUC-$2007$.}}
\begin{center}
\begin{tabular}{p{50mm}|p{9mm}|p{9mm}}
\hline
\hline
{\bf Model} & {\bf R-1} & {\bf R-2} \\
\hline
Lead \cite{acl-98:Wasson} & $0.312$ & $0.058$ \\
LexRank \cite{jair-04:Erkan} & $0.378$ & $0.075$ \\
TextRank \cite{emnlp-04:Mihalcea} & $0.403$ & $0.083$ \\
SPARSE \cite{aaai-15:Liu} & $0.354$ & $0.064$ \\
\hline
RepRank (Glove) & $0.390$ & $0.093$ \\
RepRank (Self-attention) & $0.382$ & $0.088$ \\
RepRank (Self-attention) + Absorb & $0.385$ & $0.089$ \\

\hline
\hline

\end{tabular}
\end{center}
\end{table}

The results reported in Table \ref{performance_duc_2002} also show that RepRank outperforms the other unsupervised methods for the  multi-document summarization, highlighting the potential of the proposed method based on the distributed representations. Such representations are introduced to compute the similarities between the words and sentences embedded in the same vector space, and make us possible to capture the phenomena of homonymy and polysemy in different shallow expressions. Note that all the RepRank's hyper-parameters are not tuned for any data set in the experiments. We found the similar trends from the results of $2007$ as that of DUC-$2002$, and the results for all the benchmark datasets show that RepRank achieves higher performance over the competitors.

\section{Related Work}

Many summarizers, mostly extractive, have been developed in recent years. Extractive MDS is often cast as a budgeted subset selection problem \cite{ecal-07:McDonald,acl-11:Lin} where the document collection is considered as a set of sentences, and MDS is to select a subset of the sentences under a length constraint. State-of-the-art systems solve this discrete optimization problem using integer linear programming (ILP) or submodular function maximization \cite{ilpnlp-09:Lin,ftml-12:Kulesza,acl-13:Li,eacl-14:Hong,ranlp-15:Mogren}. 

Research is moving beyond extraction in various directions: we can perform text manipulation such as compression as an important post-processing step after extraction, or alternatively, base a summary on an internal semantic representation such as the proposition \cite{acl-16:Fang}. Schluter and S{\o}gaar \shortcite{acl-15:Schluter} show that using concepts relying on syntactic dependencies or semantic frames instead of bigrams leads to significant improvements. Li et al \shortcite{naacl-15:Li} improved bigram-based ILP summarization methods by using syntactic information as well as external corpora and knowledge bases (such as Wikipedia, DBpedia, WordNet, and SentiWordNet) to select more salient bigrams.


Most extractive methods to date identify sentences based on human-engineered features. These include surface features such as sentence position and length, the words in the title, the presence of proper nouns, content features such as word frequency, event features such as action nouns, and even semantic features such as concepts and relations from external knowledge bases. 
How to determine the language features and measure their weights is the key factor impacting the system performance. This approach is effective because researchers can incorporate a large body of linguistic and domain knowledge into the models. However, this approach does not generalize well since the features carefully chosen for a domain may not be optimal to others.

Previous investigations show that human-written summaries are more abstractive, which can be regarded as a result of sentence aggregation and fusion \cite{naacl-00:Jing,acl-13:Cheung}. Bing et al \shortcite{acl-15:Bing} proposed a abstractive approach for multi-document summarization by extracting noun and verb phrases, and merging those phrases to construct new sentences. Rush et al \shortcite{emnlp-15:Rush} were the first to apply neural networks to abstractive document summarization, achieving state-of-the-art performance on the sentence-level summarization. 
However, those approaches suffer from both the limited amount of training data and the higher complexity of machine learning models.

Our approach is more close to the summarization models of \cite{acl-07:Wan,acl-16:Cheng,emnlp-17:Wang} with the differences mainly in  graph-based ranking algorithm, approximate sampling method, and distribution representation-based similarity. Wan et al \shortcite{acl-07:Wan} proposed an iterative approach for simultaneous text summarization and keyword extraction. Wang et al \shortcite{emnlp-17:Wang} formulated MDS as an affinity-preserving random walk and uses the ``global normalization'' to transform sentence affinity matrix into sentence transition matrix. 
They all use traditional tf-idf weighting method to compute the similarity between sentences and words, while we learn to embed words and sentences into a shared vector space and calculate the similarity based on their distances in such space. Besides, we prove that the iterative computation is not required for the joint task in graph-based ranking framework, and present a sample algorithm for absorbing random walks. Cheng and Lapata \shortcite{acl-16:Cheng} developed a data-driven extractive framework for single-document summarization which includes a neural network-based hierarchical document encoder (a standard recurrent neural network following a convolutional layer) and a sentence (or word) extractor with attention mechanism. Their sentence or word extractor needs to be trained separately while ours can rank the salience of sentences and words in a unsupervised manner.

\section{Conclusion}

We described an unsupervised graph-based ranking model RepRank for jointly extracting summary and keywords, in which both tasks are mutually reinforced. We proposed a word and sentence representation learning method, particularly designed for the summarization. We also proved that the preliminary iterative solution can be reduced to the absorbing random walks on a single graph consisting of both the sentences and words from an input text, which speeds up the summary generation. Experiments on multiple summarization datasets yield promising results, and RepRank can be used as a off-the-shelf tool to domains lack of training corpus 

Recent reports tend to suggest that advances in extractive document summarization have slowed down in the past few years \cite{mtd-12:Nenkova}. It seems that state-of-the-art techniques for extractive summarization have more or less achieved their peak performance and only some small improvements can be further achieved \cite{acl-16:Mehta}. This study shows that extraction summarization technique still can produce an improvement in an unsupervised way, and its optimization is not yet over, especially when the distributed representations are introduced to estimate the similarity between the language concepts for their capability in better capturing the semantic-rich  representations of different expressions. 

\section*{Acknowledgements}
This work was supported by Shanghai Municipal Science and Technology Project (No. 21511102800).

\bibliography{acl2019}
\bibliographystyle{acl_natbib}
\end{document}